\documentclass[10pt,conference]{IEEEtran}
\IEEEoverridecommandlockouts
\usepackage{cite}
\usepackage{amsmath,amssymb,amsfonts}
\usepackage{algorithmic}
\usepackage{graphicx}
\usepackage{textcomp}
\usepackage{xcolor}
\usepackage{hyperref}

\usepackage{booktabs}

\ifCLASSOPTIONcompsoc
 \usepackage[caption=false,font=normalsize,labelfont=sf,textfont=sf]{subfig}
 \else
 \usepackage{subfig}
\fi
\usepackage{algorithm}
\usepackage{algorithmic}

\floatname{algorithm}{Algorithm} %Customize to your needs
\makeatletter
\newcommand{\removelatexerror}{\let\@latex@error\@gobble}
\makeatother

\def\BibTeX{{\rm B\kern-.05em{\sc i\kern-.025em b}\kern-.08em
    T\kern-.1667em\lower.7ex\hbox{E}\kern-.125emX}}
\begin{document}

\title{GeNet: A Graph Neural Network-based Anti-noise Task-Oriented Semantic Communication Paradigm
}

\author{
Chunhang~Zheng, Kechao~Cai \\
{School of Electronics and Communication Engineering,
Sun Yat-sen University, Shenzhen, China} \\
zhengchh6@mail2.sysu.edu.cn, caikch3@mail.sysu.edu.cn}

% \author{\IEEEauthorblockN{1\textsuperscript{st} Chunhang Zheng}
%     \IEEEauthorblockA{\textit{Dept. of
%             electronics and communication engineering} \\
%         \textit{Sun Yat-sen University}\\
%         Shenzhen, China \\
%         zhengchh6@mail2.sysu.edu.cn}
%     \and
%     \IEEEauthorblockN{2\textsuperscript{nd} Caike Chao}
%     \IEEEauthorblockA{\textit{Dept. of
%             electronics and communication engineering} \\
%         \textit{Sun Yat-sen University}\\
%         Shenzhen, China \\
%         caikch3@mail.sysu.edu.cn}
% }

\maketitle

\begin{abstract}
    Traditional approaches to semantic communication tasks rely on the knowledge of
    the signal-to-noise ratio (SNR) to mitigate channel noise. Moreover, these
    methods necessitate training under specific SNR conditions, entailing
    considerable time and computational resources.
    In this paper, we propose GeNet, a Graph Neural Network (GNN)-based paradigm for
    semantic communication aimed at combating noise, thereby facilitating
    Task-Oriented Communication (TOC).
    We propose a novel approach where we first transform the input data image into graph structures.
    Then we leverage a GNN-based encoder to extract semantic information from the source data.
    This extracted semantic information is then transmitted through the channel. At the receiver's end, a GNN-based decoder is utilized to reconstruct the relevant semantic information from the source data for TOC. Through experimental evaluation, we show GeNet's effectiveness in anti-noise TOC while decoupling the SNR dependency.
    We further evaluate GeNet's performance by varying the number of nodes,
    revealing its versatility as a new paradigm for semantic communication.
    Additionally, we show GeNet's robustness to geometric transformations by testing
    it with different rotation angles, without resorting to data augmentation.

    % In this paper, we propose a Graph Neural Network (GNN)-based anti-noise semantic
    % communication paradigm, named GeNet to facilitate Task-Oriented Communication
    % (TOC). 
    % In semantic communication tasks, existing work relies on the knowledge of
    % signal-to-noise ratio (SNR) to deal with noise in the channel. 
    % Specifically, these models must be trained under specific SNR conditions, which
    % spend significant time and computational resources. 
    % To address this issue, we transform the input data image into graph structures
    % and then use GNN-based encoder to extract the semantic information from the
    % source data. 
    % The extracted semantic information is then transmitted through the channel. 
    % At the receiver's end, we use GNN-based decoder to reconstruct the interested
    % semantic information of source data for TOC. 
    % We demonstrate the effectiveness of GeNet in anti-noise TOC while decoupling the
    % SNR. 
    % We also set different number of nodes to evaluate the performance of GeNet,
    % which indicates that our model can be applied as a new paradigm for semantic
    % communication. 
    % Additionally, the model is tested with different rotation angles to show its
    % robustness to geometric transformations without any data augmentation.

\end{abstract}


% ICCCN
% 1. abstract 300 words
% 2. introduction 1.1 页
%   - important
%   - 别人的缺陷
%   - Framework的优势
% 3 related work 0.5
% 4. Framework, Model 3
% 5. Experiments 3.5
% 6. Conclusion 0.3
% 7. reference 20 以上

\section{Introduction}

% the importance of semantic communication
In an era where traditional communication technologies approach the Shannon
limit, the emergence of Artificial Intelligence (AI) has fostered innovative
paradigms such as Semantic Communication (SC).
Unlike conventional methods that solely focus on accurately transmitting bits,
SC delves into the core of transmitted information, aiming to preserve its
meaning across the communication channel~\cite{sc2}.
This approach is particularly crucial in environments with high network
complexity and limited spectrum resources, as it significantly reduces data
transmission load, thereby enhancing communication efficiency.
Among the various aspects of SC, Task-Oriented Communication (TOC) distinguishes
itself by emphasizing the extraction of receiver-relevant semantic information
from the source~\cite{sc1}, thereby enabling more informed and effective task
inference at the receiver's end.

% In the era where traditional communication technologies are inching closer to the Shannon limit, the advent of Artificial Intelligence (AI) has paved the way for innovative paradigms such as Semantic Communication (SC). Unlike traditional methods focusing merely on the accurate transmission of bits, SC delves into the essence of transmitted information, aiming to preserve the meaning behind the data across the communication channel~\cite{sc2}. This approach is particularly vital in scenarios with high network complexity and scarce quality spectrum resources, as it substantially reduces data transmission load, thereby enhancing communication efficiency. Among various facets of SC, Task-Oriented Communication (TOC) stands out by prioritizing the extraction of receiver-relevant semantic information from the source~\cite{sc1}, facilitating more informed and effective task inference at the receiver's end.
% \red{added}
SC has been widely studied in the literature, with most previous methods relying on neural networks (NN) to extract semantic information from the source data and reconstruct it at the receiver's end. As an aspect of SC, TOC provides a unique perspective by focusing on the ultimate goal or task for which the communication is intended, rather than merely ensuring the fidelity of information transmission. This shift towards task-oriented communication encourages the development of systems that are not only efficient in terms of bandwidth usage but also in terms of the computational resources required for processing the transmitted data.

\begin{figure}[t!]
	\centering
	\includegraphics[width=1.0\linewidth]{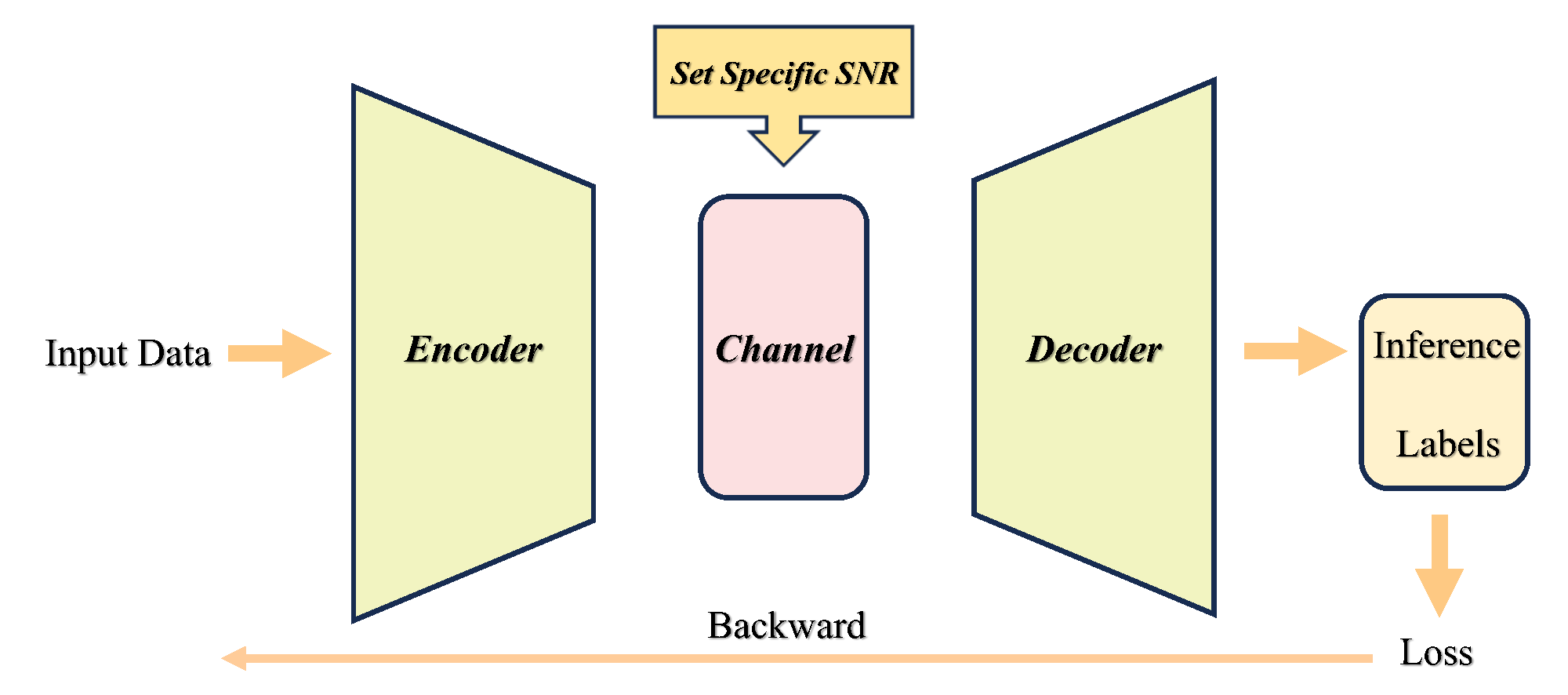}
	\caption{General semantic communication paradigm. Previous methods need to \emph{set specific SNR}.}
	\label{conv}
\end{figure}

However, all previous SC methods for noise handling in channels rely on the knowledge of the SNR. Specifically, these models must be trained under specific SNR conditions,
shown in Fig.~\ref{conv}. This paradigm of
non-decoupling the SNR in channel
from the SC progress could result in two undesirable phenomean: 1) the model trained with high SNR performs worse in low SNR, compared with the model orginally trained with low SNR; 2) the model trained with low SNR does not improve significantly when the channel condtion i.e., SNR, is improved.
% and their performance may degrade significantly when operating under different SNR conditions than those encountered during training. 
In this case, models should be trained in different channel conditions to get a satisfied performance, which requires significant time and computational resources.

% the emergence of graph neural network
% On the other hand,

Moreover, Graph Neural Network (GNN) has emerged as a powerful tool for
processing graph-structured data, and it has been widely applied in various fields, including recommendation systems \cite{recommendation},
drug discovery \cite{drug}, and computer vision.
The success of GNN in these fields has inspired researchers to explore its
potential in wireless communications.
In particular, GNN has been applied to wireless communication problems such as
task offloading \cite{offloading} and cellular traffic prediction \cite{cellular}.
% \red{add citation}

% The success of GNN in these problems has demonstrated the potential of GNN in
% wireless communications.
% \red{add paragraph}

% htbp

% our method and contribution
In this paper, we explore a new paradigm for SC, named GeNet, which leverages
GNN to facilitate TOC. GeNet is designed to address the challenges of anti-noise
TOC, where the transmitted data in a channel is corrupted by noise while decoupling the SNR.
Specifically,
GeNet comprises a GNN-based encoder and a GNN-based decoder. The GNN-based
encoder is designed to extract semantic information from the source data, while
the GNN-based decoder is designed to reconstruct the task-relevant semantic
information of source data for TOC in downstream. We demonstrate the
effectiveness of GeNet in anti-noise TOC through extensive experiments on
MNIST~\cite{mnist}, FashionMNIST\cite{fashionmnist} and CIFAR10~\cite{cifar10}
datasets. Additionally, our source code has been released and is publicly
available\footnote[1]{\url{https://github.com/chunbaobao/GeNet}}.

% specific contributions
In summary, our work makes the following contributions:
\begin{itemize}
	\item To the best of our knowledge, we are the first to apply GNN to anti-noise
	      semantic communication. We propose a novel GNN-based framework for anti-noise
	      semantic communication, which consists of a GNN-based encoder and a GNN-based
	      decoder, and explored the possibility of it serving as a semantic communication
	      paradigm.
	\item Our model can denoise on the AWGN channel without knowing the SNR. Specifically, unlike other anti-noise semantic communication methods, our method is able to perform anti-noise semantic communication while decoupling the SNR. We demonstrate the effectiveness of GeNet in anti-noise TOC through extensive experiments on MNIST, FashionMNIST and CIFAR10 datasets.
	\item Compared with traditional CNN-based methods, our GNN-based model can process images of different pixels without resizing the image that may cause information loss or redesigning and retraining the model architecture. This is crucial in scenarios where the image size is not fixed or the image size is large.
	\item Additionally, our model can extract features that are inherently invariant to geometric transformations such as rotations ensuring that the essence of the transmitted information is preserved and accurately interpreted at the receiving end without data augmentation.
\end{itemize}

% organization of paper
The rest of this paper is organized as follows. Section \ref{RW} introduces the related work on DL-based semantic communication and graph neural network in wireless communications. Section \ref{GeNet} presents the preprocessing of image data and the proposed GeNet framework. In Section \ref{Experiments}, we provide extensive simulation results to evaluate the performance of the proposed methods. Finally, Section \ref{Conclusion} concludes the paper.

\section{Related Work}\label{RW}

\subsection{DL-Based Semantic Communication}

The purpose of the next generation of communication systems has evolved from reliable transmission of bit sequences to a wider range of objectives, namely
providing intelligent, efficient, and sustainable communication services. Its core is to extract the message semantics of the transmitting end based on the task requirements of the receiving end, and
to achieve more intelligent communication and interaction based on the context and requirements of the receiving end, aiming to interpret the meaning of the transmitted information accurately across the communication channel to improve the efficiency of communication systems.

DL techniques have shown great potential in processing various intelligent tasks, i.e., computer vision and NLP~\cite{sc4}. In the communication area, the recent line of research on semantic communication~\cite{sc5,djscc,djscc2,muti_edeg} has motivated a paradigm shift in communication system design for many types of data.

In particular, a task-oriented semantic communication system was proposed in~\cite{sc5} that provides text and speech of the target language at the receiving end by transmitting low dimensional semantic features to achieve translation tasks in speech transmission, and exhibits higher robustness in the case of channel damage. A task-oriented communication for multi-device cooperative edge inference was proposed in~\cite{muti_edeg}, where a group of low-end edge devices transmit the task-relevant features to an edge server for aggregation and processing.
In the seminal work of semantic communication~\cite{djscc}, E.~Bourtsoulatze et al.\ construct an encoder-decoder architecture through neural networks and uses the energy constrained encoder outputs.
The output of the beam encoder is transmitted through AWGN and Rayleigh fading channels, achieving better performance in noisy environments than traditional image coding JPEG and JPEG2000.

% The output of the beam encoder is transmitted through AWGN and Rayleigh fading channels, achieving comparative transmission in noisy environments
% Better performance of unified image encoding JPEG and JPEG2000.

% However, all existing methods for noise handling in channels rely on knowledge of the SNR. Specifically, these models must be trained under specific SNR conditions, shown in Fig.~\ref{conv}, and their performance may degrade significantly when operating under different SNR conditions than those encountered during training. In this case, models should be trained in different channel condition to get a satisfied performance, which need significant time and computational resources.

\subsection{Graph Neural Network in Wireless Communication}

% With the widespread application of graph data, GNN has become one of the hotspots in research. Their learning capability on graph-structured data has demonstrated immense potential across various applications. GNNs originated from~\cite{gft}, which extended the Fourier transform to graph structures, namely the Graph Fourier Transform, laying the foundation for spectral graph convolution networks. In foundational GNN research, spectral-based Graph Convolutional Network (GCN) was proposed based on the Graph Fourier Transform~\cite{gcn}, attention mechanism was utilized in~\cite{gat} to autonomously learn the importance of edge relationships between nodes. In applied GNN research,~\cite{apgnn2} proposed a method for dynamic WLAN performance prediction using a heterogeneous temporal graph neural network, while~\cite{apgnn1} utilized GNNs and GLUs to extract spatial and temporal information, respectively, to accomplish traffic prediction tasks in cellular networks.

GNN have emerged as a focal point in research due to their remarkable learning capability on graph-structured data across diverse applications.
As an extension of neural networks, GNN can handle data formats represented by
graph structures. In the graph, each node is defined by its own features and its
adjacent nodes and relationships, and the network calculates the node's
representation vector by recursively aggregating and transforming the
representation vectors of adjacent nodes. The goal of GNN is to learn useful
representations from graph structured data and use these representations for
various tasks, such as node classification, graph classification, link
prediction, and so on.

GNN is originated from seminal work by~\cite{gft}, which introduced the Graph
Fourier Transform, extending the Fourier transform to graph structures and
paving the way for spectral graph convolution networks.
Foundational research in GNN includes the introduction of spectral-based Graph
Convolutional Network (GCN) by~\cite{gcn}, leveraging the Graph Fourier
Transform.
Furthermore, attention mechanisms, as seen in~\cite{gat}, autonomously learn the
importance of edge relationships between nodes.
In applied GNN research,~\cite{apgnn2} proposed a method for dynamic WLAN
performance prediction using a heterogeneous temporal graph neural network,
while~\cite{apgnn1} employed GNN and GLUs to extract spatial and temporal
information, respectively, for traffic prediction tasks in cellular networks.

\section{GNN-Based SC Framework}\label{GeNet}

\begin{figure*}[t!]
	\centering
	\includegraphics[width=0.9\linewidth]{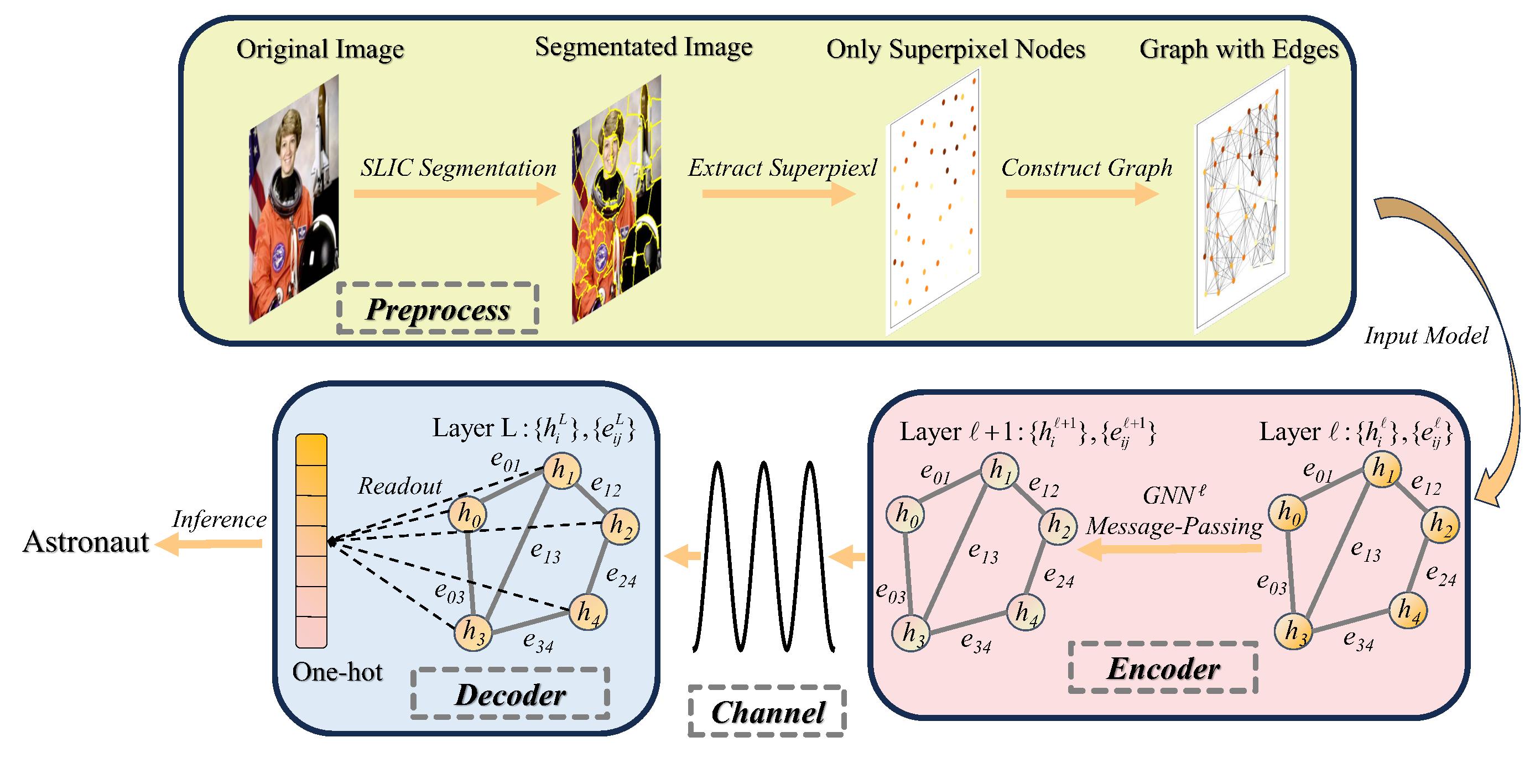}
	\caption{Our GeNet model architecture.}
	\label{pipeline}
\end{figure*}

Most models for semantic communication, specifically task-oriented communication, have an encoder-decoder structure~\cite{djscc2,djscc,muti_edeg}. Here, the input data representations $(x_{1},...,x_{n})$ are mapped to a low-dimensional semantic space $(z_{1},...,z_{m})$ by the encoder, and the decoder maps the low-dimensional semantic information to the data representations $(\hat{y}_{1},...,\hat{y}_{o})$ which is relevant to the receiver for downstream task. The encoder and decoder are trained to minimize the error between the inference and task-relevant labels $({y}_{1},...,{y}_{o})$.

Our proposed GeNet framework follows this overall architecture that consists of a GNN-based encoder and a GNN-based decoder, shown in the Fig.~\ref{pipeline}. The details of the GeNet model are described in the following sections.

\subsection{Preprocessing}\label{Preprocess}
The preprocessing comprises two primary steps. First, we extract the segmentation of the input image. Second, we transform the segmentation information into the graph structure.

In the first step, we use SLIC segmentation method \cite{slic,slicclassfications,extract_superpixel} to segment the popular MNIST, FashionMNIST and CIFAR10 image classification datasets. SLIC is a simple and efficient method for image segmentation, which is based on the K-means clustering algorithm. It is able to generate superpixels that are more regular in shape and size, and it is also able to preserve the boundaries of the objects in the image.

In the second step, we transform the segmentation into a graph structure similar to \cite{benchmark}. Specifically, we use the superpixels as the nodes of the graph, and the features of the nodes are the average color and the centroid of the superpixels.
The edges are connected between the nodes and their $k$-th nearest neighbors, and the weight of the edge is determined by the similarity of the nodes which is calculated by Euclidean distance between the the features of the nodes:

\begin{align}
	 & \mathbf{A}_{ij}^{k}=\exp\left(-\frac{\|\boldsymbol{f}_i-\boldsymbol{f}_j\|^2}{\sigma_{i}^2}\right),\label{A} \\
	 & \sigma_i = \sum_{1}^{k} \|\mathcal{D}_i^{k}\|^2 +\epsilon,
	\label{sigma}
\end{align}
where $\boldsymbol{f}_i$ and $\boldsymbol{f}_j$ are the features of the superpixel nodes $i$ and $j$, $\epsilon$ is a small constant to avoid division by zero, $\sigma_i$ is the scale parameter, and the $\mathcal{D}_i^{k}$ is the similarity between the node $i$ and its $k$-th nearest neighbor.

The superpixel graph generation procedures from image are summarized in Algorithm \ref{algorithm:1}. In this algorithm, we first compute the features of each nodes that are the average color and the centroid of the superpixels. We then construct the adjacency matrix based on Euclidean distance between the the features of the nodes.

% we first compute the color and spatial distance of the superpixels, and then we construct the adjacency matrix based on the color and spatial distance between the superpixels. The features of the nodes in graph are the average color and the centroid of the superpixels. And weight of the edge is determined by the similarity of the superpixel nodes.

\begin{figure}[!t]

	\removelatexerror
	\begin{algorithm}[H]
		\caption{Superpixel Graph Generation from Segmentation}
		\label{algorithm:1}
		\begin{algorithmic}[1]
			\REQUIRE Origial Image $\mathbf{I}$, segmentation mask $\mathbf{S}$, number of nodes $N$.
			\ENSURE Superpixel Graph $\mathcal{G} = (\mathcal{N}, \mathcal{E})$.
			\STATE \textbf{Initialization:} Set $\mathcal{N} \gets \emptyset$, $\mathcal{E} \gets \emptyset$.

			\STATE \textbf{for} $i=1$ to $N$ \textbf{do}
			\STATE \quad Compute color feature $\boldsymbol{c}_i$ as the average color of superpixel $\mathbf{S}(i)$ in $\mathbf{I}$.
			\STATE \quad Compute coordinate feature $\boldsymbol{p}_i$ as the centroid of superpixel $\mathbf{S}(i)$ in $\mathbf{I}$.
			\STATE \quad Add node $i$ to $\mathcal{N}$ with features $\boldsymbol{f}_i = \left[\boldsymbol{c}_i, \boldsymbol{p}_i\right]$.
			\STATE \textbf{end for}
			\STATE Computer adjacency matrix $\mathbf{A}$ based on \eqref{A} and \eqref{sigma}.
			\STATE \textbf{return:} Graph $\mathcal{G} = (\mathcal{N}, \mathcal{E})$ with adjacency matrix $\mathbf{A}$.
		\end{algorithmic}
	\end{algorithm}
\end{figure}

\subsection{Encoder and Decoder}\label{Ed}
There are two representive GNN frameworks: Message Passing Neural Network (MPNN) \cite{mpnn} and Weisfeiler Lehman Network GNNs (WL-GNNs). MPNNs operate by passing messages between nodes in a graph, allowing nodes to aggregate and update information from neighboring nodes iteratively. They are versatile and widely used in tasks such as graph classification, node classification, and link prediction. WL-GNNs, on the other hand, are based on the Weisfeiler-Lehman (WL) graph isomorphism test, which is a powerful tool for distinguishing non-isomorphic graphs.

In this work, we use the MPNN architecture as the backbone of our GeNet model. The MPNN architecture consists of two main components: message passing and readout. The message passing component is used to update the node features based on the features of the neighboring nodes:
\begin{align}
	 & \boldsymbol{m}_e^{(\ell+1)}=\phi\left(\boldsymbol{h}_v^{(\ell)},\boldsymbol{h}_u^{(\ell)},\boldsymbol{w}_e^{(\ell)}\right),(u,v,e)\in\mathcal{G}, \label{message}        \\
	 & \boldsymbol{h}_v^{(\ell+1)}=\psi\left(\boldsymbol{h}_v^{(\ell)},\rho\left(\left\{\boldsymbol{m}_e^{(\ell+1)}:(u,v,e)\in\mathcal{G}\right\}\right)\right), \label{update}
\end{align}
where $\boldsymbol{h}_v^{(\ell)}$ and $\boldsymbol{h}_u^{(\ell)}$ are the features of the nodes $v$ and $u$ at the $\ell$-th layers, respectively, and $\boldsymbol{w}_e^{(\ell)}$ is the feature of the edge $e$. We denote $\phi$ as the message function defined for every edge in the graph.  $\rho$ denotes the aggregation function that aggregates the messages received by nodes; $\psi$ denotes the update function that updates the node features based on the aggregated messages and the node features itself.

The readout component is used to aggregate the node features in the last layer to obtain the graph-level features:
\begin{equation} \label{readout}
	\boldsymbol{h}_\mathcal{G}=\rho_\mathcal{G}\left(\left\{\boldsymbol{h}_v^{(L)}:v\in\mathcal{N}\right\}\right),
\end{equation}
where $\boldsymbol{h}_\mathcal{G}$ is the graph-level features, $L$ is the number of stacked GNN layers, and $\rho_\mathcal{G}$ is the readout function to obtain the graph-level features. The graph-level features are then used to make the final prediction.

In our GeNet model, the encoder and decoder correspond to the message passing and readout components of the MPNN architecture, respectively. The encoder is used to extract the semantic information from the source data based on~\eqref{message} and~\eqref{update}, and the decoder is used to reconstruct the interesting semantic information of source data for TOC based on~\eqref{readout}.

As for the channel, we use the AWGN channel to simulate the noise in the
communication.
We add  noise to node features using Algorithm~\ref{algorithm:2}.
In this algorithm, we first compute the channel dimension and the signal power
of the graph, and then we compute the noise power based on the SNR, which is
defined as follows:
\begin{equation} \label{eq:snr}
	\mathrm{SNR}=10\log_{10}\frac{P}{\sigma^2}\quad\mathrm{(dB)}.
\end{equation}

We then sample the noise from the Gaussian distribution and add the noise to the node features. The noisy node features are then used as the input to the decoder.
It is worth noting that since the decoder directly performs readout processing on node features to obtain graph-level features, there is no need to simulate noise addition to the edges of the graph in the channel.

\begin{figure}[!t]
	\removelatexerror
	\begin{algorithm}[H]
		\caption{Add Noise to Node Features}
		\label{algorithm:2}
		\begin{algorithmic}[1]
			\REQUIRE Features matrix of nodes $\mathbf{H}\in\mathbb{R}^{N \times d}$, noise level $\mathrm{SNR}$ (dB).
			\ENSURE Noisy features matrix of nodes $\mathbf{H}'$.
			\STATE Compute the channel dimension $k = N \times d$.
			\STATE Compute the signal power $P = \frac{\sum_{i=1}^{N} \sum_{j=1}^{d} \mathbf{H}_{ij}^2}{k} $.

			\STATE Compute noise power $\sigma_n^2$ based on \eqref{eq:snr}.
			\STATE \textbf{for} $v=1$ to $N$ \textbf{do}
			\STATE \quad Sample noise $\boldsymbol{n}_v\sim\mathcal{N}(0,\sigma_n^2)$ with $\boldsymbol{n}_v\in\mathbb{R}^{d}$.
			\STATE \quad Add noise to node feature $v$ $\boldsymbol{h}'_v = \boldsymbol{h}_v + \boldsymbol{n}_v$.
			\STATE \textbf{end for}
			\STATE \textbf{return:} Noisy features matrix of nodes $\mathbf{H}'$.
		\end{algorithmic}
	\end{algorithm}
\end{figure}

% why choose mean readout
In our GeNet model, we use the mean readout function to aggregate the node features in the decoder to obtain the graph-level features, which can reduce the noise in the channel. The mean readout function is defined as:
\begin{equation}\label{mean}
	\boldsymbol{h}_\mathcal{G} = \frac{1}{N} \sum_{v \in \mathcal{N}} \boldsymbol{h}_v^{(L)}.
\end{equation}

Assuming that the noise is independent and identically distributed (i.i.d.). As mentioned in Algorithm \ref{algorithm:2}, the noisy feature of node $v$ is $\boldsymbol{h}'_v = \boldsymbol{h}_v + \boldsymbol{n}_v$. Given that the mean readout function, we have the following graph-level features when the noise is added to the node features:

\begin{equation} \label{eq:mean_snr}
	\begin{aligned}
		\boldsymbol{h}_\mathcal{G} & = \frac{1}{N} \sum_{v \in \mathcal{N}} \left(\boldsymbol{h}_v^{(L)} +  \boldsymbol{n}_v \right)                           \\
		                           & =   \frac{1}{N} \sum_{v \in \mathcal{N}} \boldsymbol{h}_v^{(L)}  + \frac{1}{N} \sum_{v \in \mathcal{N}} \boldsymbol{n}_v.
	\end{aligned}
\end{equation}
Among them, $\boldsymbol{n}_v\sim\mathcal{N}(0,\sigma_n^2)$, and $\frac{1}{N} \sum_{v \in \mathcal{N}} \boldsymbol{n}_v \sim\mathcal{N}(0,\frac{\sigma_n^2}{N})$. Therefore, the noise power in the graph-level features is reduced by a factor of $\frac{1}{N}$ compared to the noise power in the channel.

\section{Experiments}\label{Experiments}

In this section, we assess the performance of the proposed GeNet model on the MNIST \cite{mnist}, FashionMNIST \cite{fashionmnist} and CIFAR10 \cite{cifar10} datasets. And all experiments are conducted either on a computer equipped with a single NVIDIA GeForce RTX 3090 GPU and an Intel Core i7-12700K or one with two NVIDIA GeForce RTX 3090 GPUs and an Intel Xeon Gold 6230. For the development, we use the PyTorch \cite{pytorch} and Deep Graph Library (DGL) \cite{dgl}.

\subsection{Experiment Setup}
\subsubsection{Preprocessing}
The MNIST dataset comprises 60,000 training images and 10,000 test images of handwritten digits for 10 classes, ranging from 0 to 9. Each image is grayscale and has a resolution of 28x28 pixels.
The FashionMNIST dataset consisting of 60,000 training images and 10,000 test images of 10 classes of fashion items, shares the same resolution as MNIST.
On the other hand, the CIFAR10 dataset containing 50,000 training images and 10,000 test images of 10 classes, features RGB images with a resolution of 32x32 pixels.

To create a validation set, 10\% of each class from the training dataset is randomly selected for all datasets. It should be noted that the `6' and `9' classes in MNIST are excluded from the validation and test datasets as their rotated versions are identical. Following the approach in previous works \cite{extract_superpixel,benchmark}, specific parameter settings are applied: 95 nodes and a compactness parameter of 0.25 for MNIST, 105 nodes and a compactness parameter of 0.3 for FashionMNIST, and 200 nodes with a compactness parameter of 10 for CIFAR10.

\begin{figure}[t!]
	\centering
	\subfloat[An example from MNIST with label `3'.]{\includegraphics[width=1\linewidth]{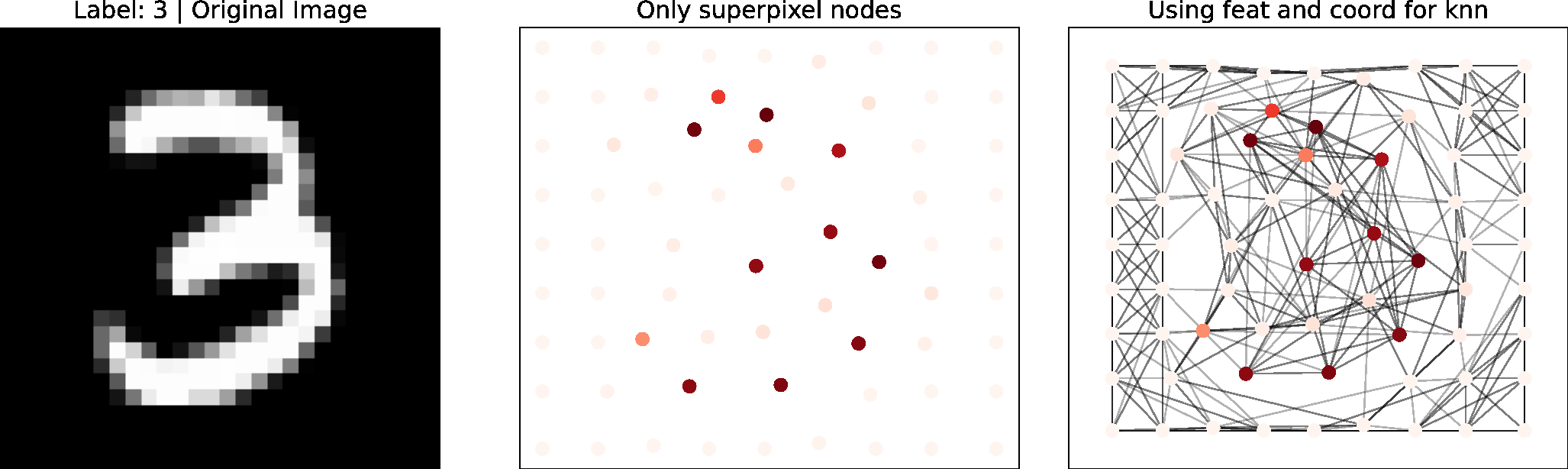}}
	\\
	\subfloat[An example from FashionMNIST with label `T-shirt'.]{\includegraphics[width=1\linewidth]{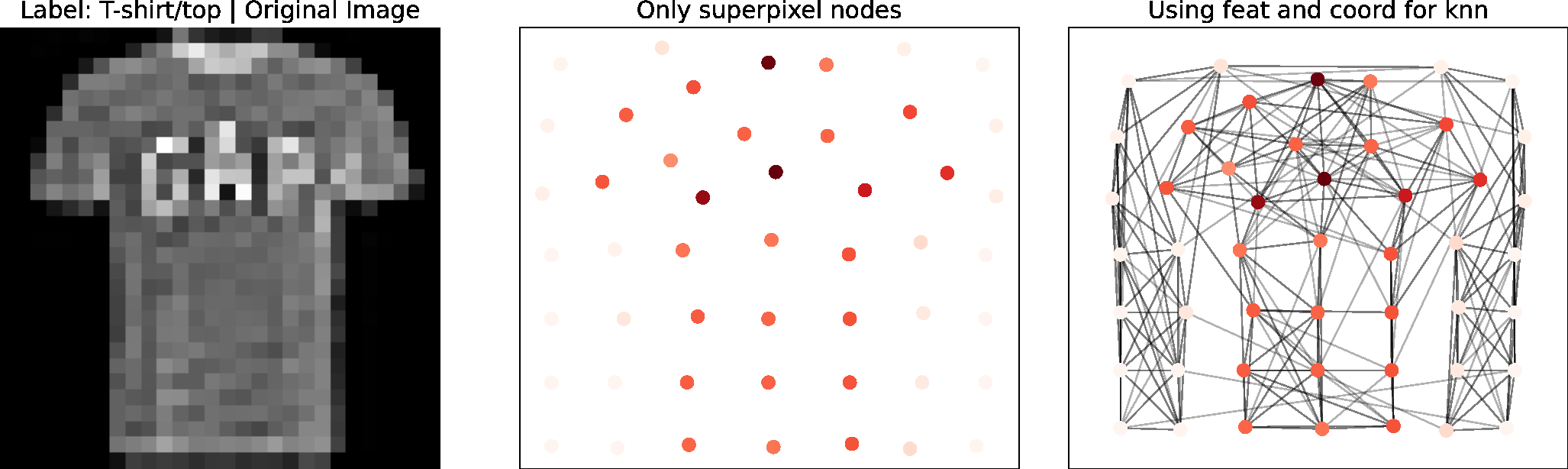}}
	\\
	\subfloat[An example from CIFAR10 with label `truck'.]{\includegraphics[width=1\linewidth]{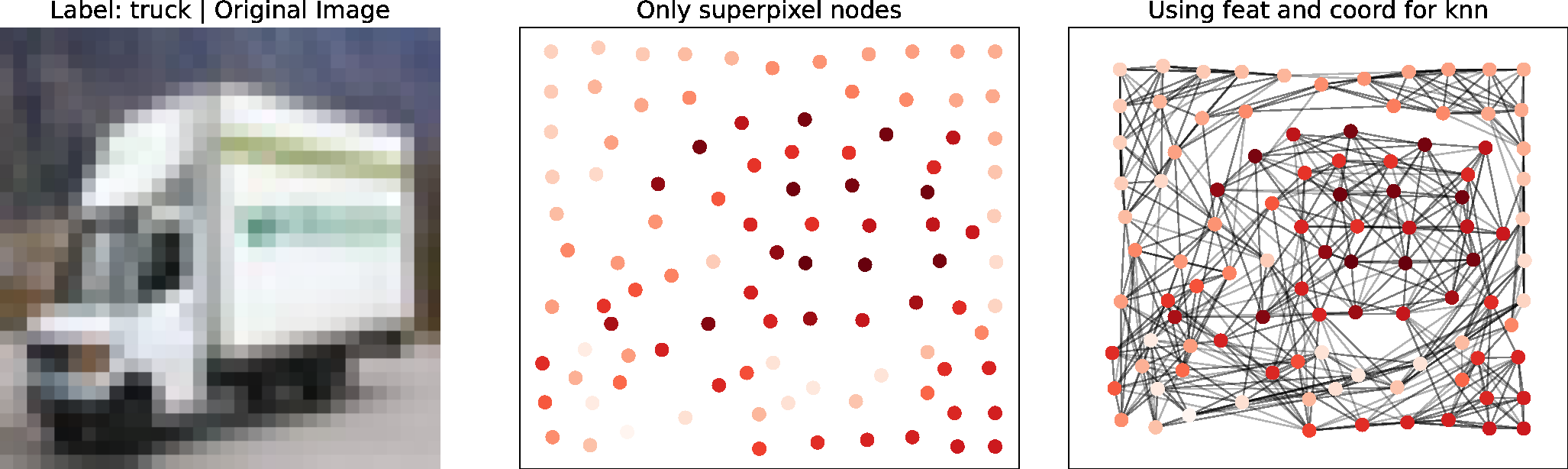}}

	\caption{Examples of the superpixel graph generation.}
	\label{example}
\end{figure}

% \red{added}
% Here are some examples of the superpixel graph generation from the image in Fig.~\ref{example}. The left sub image is the original image, the mid sub image is the superpixel nodes generated by SLIC that includes the average color information and the centroid position information, and the right sub image is the superpixel graph whose adjacency matrix is generated by Algorithm~\ref{algorithm:1}.
Furthermore, exemplifying the superpixel graph generation process from the image in Fig.~\ref{example}, the left sub-image represents the original image while the middle sub-image illustrates the superpixel nodes generated by SLIC, incorporating average color and centroid position information. The right sub-image showcases the superpixel graph with its adjacency matrix generated using Algorithm~\ref{algorithm:1}.

\begin{figure}[t!]
	\centering
	\subfloat[An example from MNIST with label `7'.]{\includegraphics[width=0.37\textwidth]{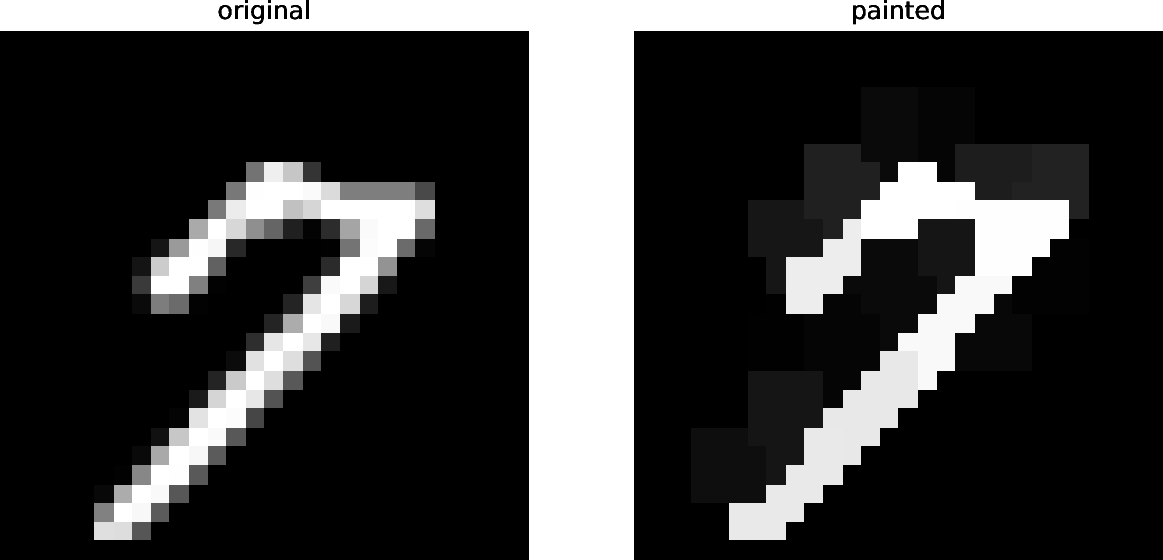}}
	\\
	\subfloat[An example from FashionMNIST with label `sneaker'.]{\includegraphics[width=0.37\textwidth]{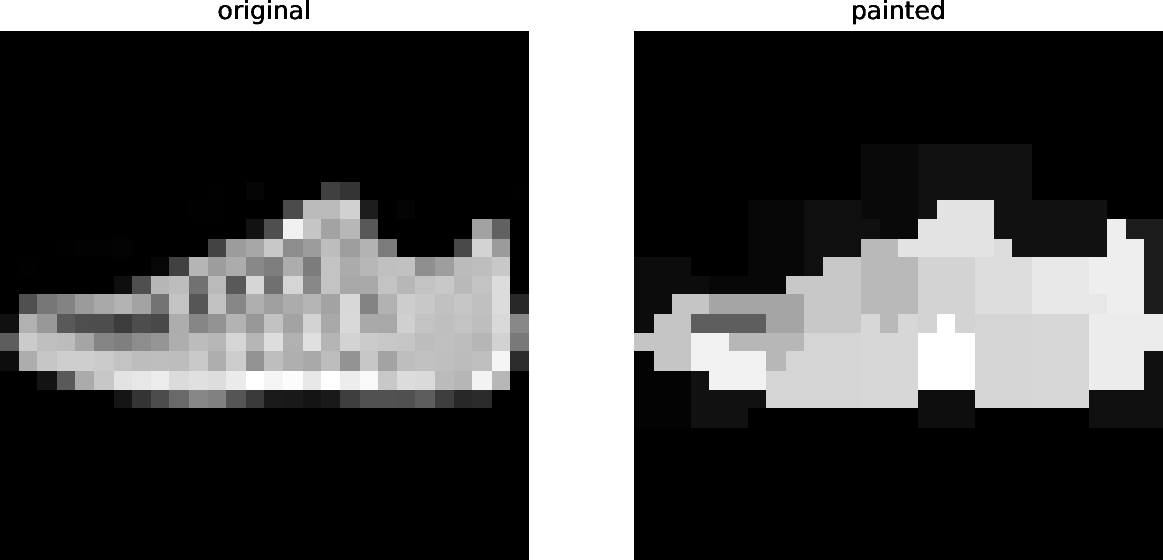}}
	\\
	\subfloat[An example from CIFAR10 with label `truck'.]{\includegraphics[width=0.37\textwidth]{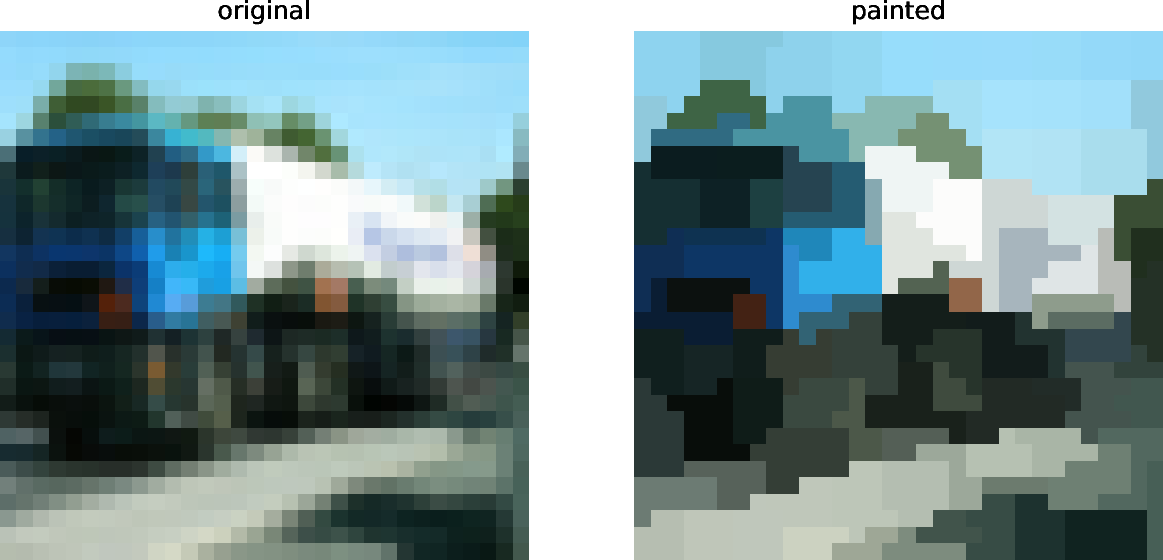}}
	\caption{Examples showing the loss of color information.}
	\label{paint}
\end{figure}

\subsubsection{Training}\label{training}

To fully demonstrate the effectiveness of our proposed model, we used various GNN models as the backbone of the GeNet model, including Graph Convolutional Network (GCN) \cite{gcn}, Graph Attention Network (GAT) \cite{gat}, Gated Graph Convolutional Network (GATEDGCN) \cite{gatedgcn}, along with GNNs augmented with Multi-layer Perceptrons (MLP). MLP does not directly use the original image as input, but uses node features converted to a graph structure as input.

The models were all configured with a consistent architecture comprising 4 stacked GNN layers. Training was conducted using the Adam optimizer \cite{adam} with an initial learning rate, which was dynamically adjusted by reducing it when the validation set accuracy failed to improve over consecutive epochs. Training ceased once the learning rate reached a specified threshold to avoid overfitting. To ensure that the model was not underfit, an ample number of training epochs were established. The cross-entropy loss function was employed to quantify the disparity between the predicted outcomes and the relevant task labels. Further elaboration on the model structures and training setups is available in Table \ref{table:config}.

% All the number of stacked GNN layers is set to 4, and all the models are trained with the Adam optimizer \cite{adam} with a initial learning rate, and the learning rate is reduced by a factor when the validation set accuracy does not improve for some epochs. When the learning rate drops to a certain value, the training will stop. To prevent underfitting of the model, all training epochs are set to a sufficiently large number. We use the cross-entropy loss to measure the difference between the predicted results and the task-relevant labels. More details about the structures of models and training configurations are summarized in Table \ref{table:config}.

\begin{table}[t!]

	\renewcommand{\arraystretch}{1.3}
	\caption{Parameters of models and training configurations.}
	\label{table:config}
	\centering
	\begin{tabular}{ccccc}
		\toprule
		                      & GatedGCN & GCN  & GAT  & MLP  \\           \midrule
		Batch Size            & 5        & 5    & 50   & 5    \\
		Initial Learning Rate & 5e-5     & 5e-5 & 5e-5 & 5e-4 \\
		LR Reduce Factor      & 0.5      & 0.5  & 0.5  & 0.5  \\
		LR Schedule Patience  & 25       & 25   & 25   & 10   \\
		Min LR                & 1e-6     & 1e-6 & 1e-6 & 1e-5 \\
		Number of Layer       & 4        & 4    & 4    & 4    \\
		Hidden   Dim          & 70       & 146  & 19   & 150  \\
		Output   Dim          & 70       & 146  & 152  & 150  \\
		Readout               & mean     & mean & mean & mean \\
		Self   Loop           & No       & No   & No   & No   \\
		Number   of Heads     & -        & -    & 8    & -    \\
		\bottomrule
	\end{tabular}
\end{table}

%%% add fashionmnist and change 2*3 in two columns to 3*2 in one column

% ! too wide for two columns 2*3
% ! consider using one column: 3*2
% \begin{figure*}[t!]
%     \centering
%     \subfloat[]{\includegraphics[width=0.3\textwidth]{figure/original_mnist_9788.eps}}
%     \hfill
%     \subfloat[]{\includegraphics[width=0.3\textwidth]{figure/original_fashionmnist_22426.eps}}
%     \hfill
%     \subfloat[]{\includegraphics[width=0.3\textwidth]{figure/original_cifar10_7622.eps}}
%     \\
%     \subfloat[]{\includegraphics[width=0.3\textwidth]{figure/paint_mnist_9788.eps}}
%     \hfill
%     \subfloat[]{\includegraphics[width=0.3\textwidth]{figure/paint_fashionmnist_22426.eps}}
%     \hfill
%     \subfloat[]{\includegraphics[width=0.3\textwidth]{figure/paint_cifar10_7622.eps}}

%     \caption{Examples showing the loss of colour information.}
%     \label{paint}
% \end{figure*}

% \begin{figure*}[t!]
%     \centering
%     \subfloat[painted example from MNIST with label '7']{\includegraphics[width=0.45\textwidth]{figure/paint_mnist_9788.eps}}\hspace{15pt}
%     \subfloat[painted example from CIFAR with label 'truck']{\includegraphics[width=0.45\textwidth]{figure/paint_cifar10_7622.eps}}
%     \caption{Examples showing the loss of colour information.}
%     \label{paint}
% \end{figure*}

% -------------------------------------------------------

% !add fashionmnist also change 1*4 to 2*3
\begin{figure*}[t!]
	\centering
	\subfloat[Train loss in MNIST.]{\includegraphics[width=0.3\textwidth]{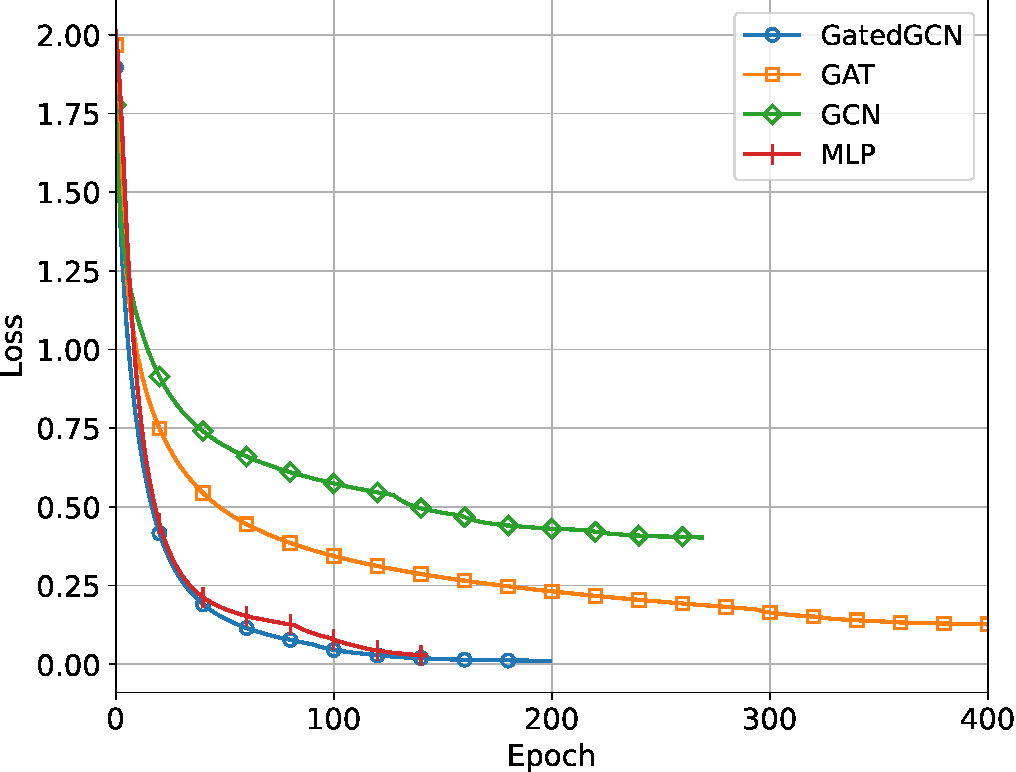}}
	\hspace{5pt}
	\subfloat[Train loss in FashionMNIST.]{\includegraphics[width=0.3\textwidth]{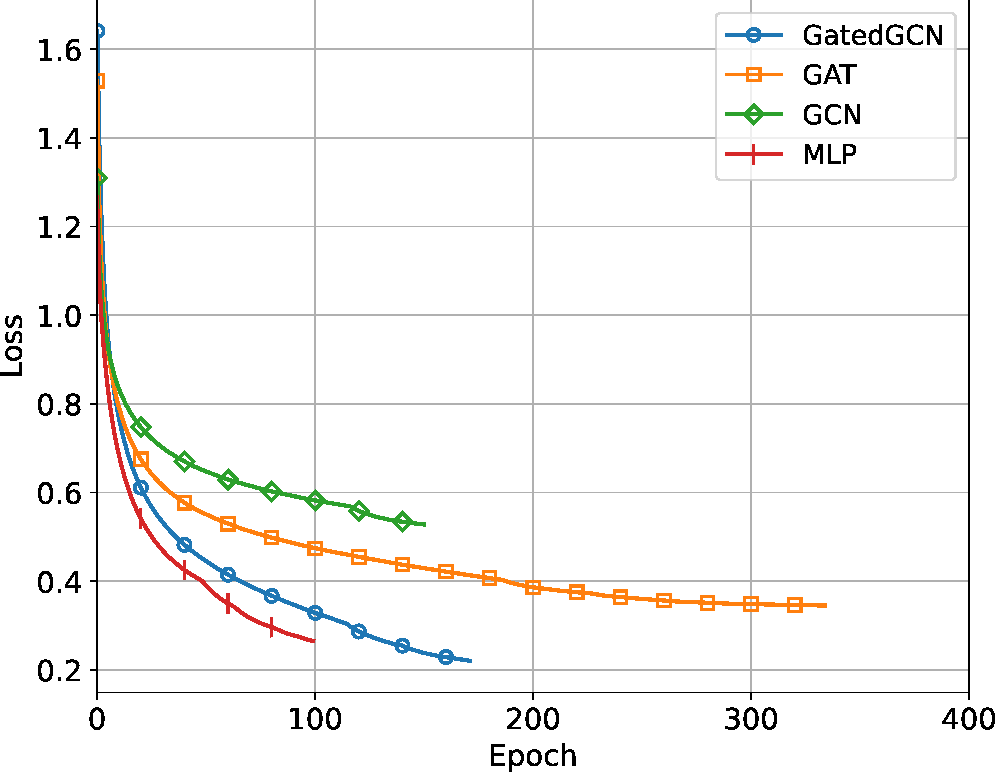}}
	\hspace{5pt}
	\subfloat[Train loss in CIFAR10.]{\includegraphics[width=0.3\textwidth]{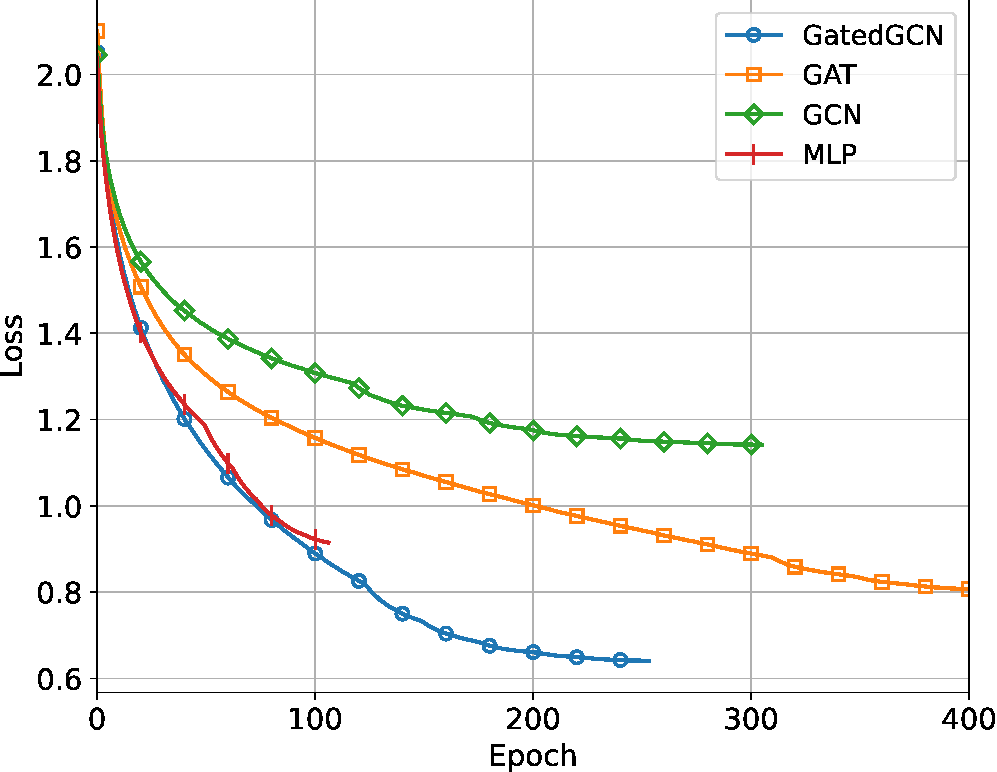}}
	\\
	\subfloat[Validation accuracy in MNIST.]{\includegraphics[width=0.3\textwidth]{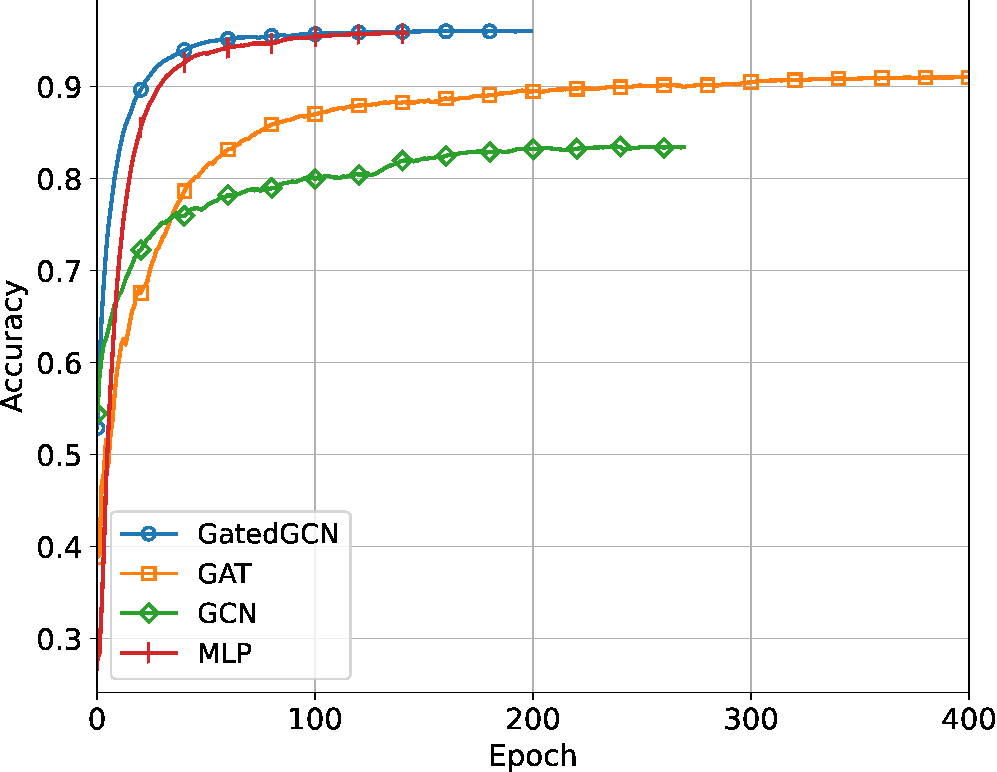}}
	\hspace{5pt}
	\subfloat[Validation accuracy in FashionMNIST.]{\includegraphics[width=0.3\textwidth]{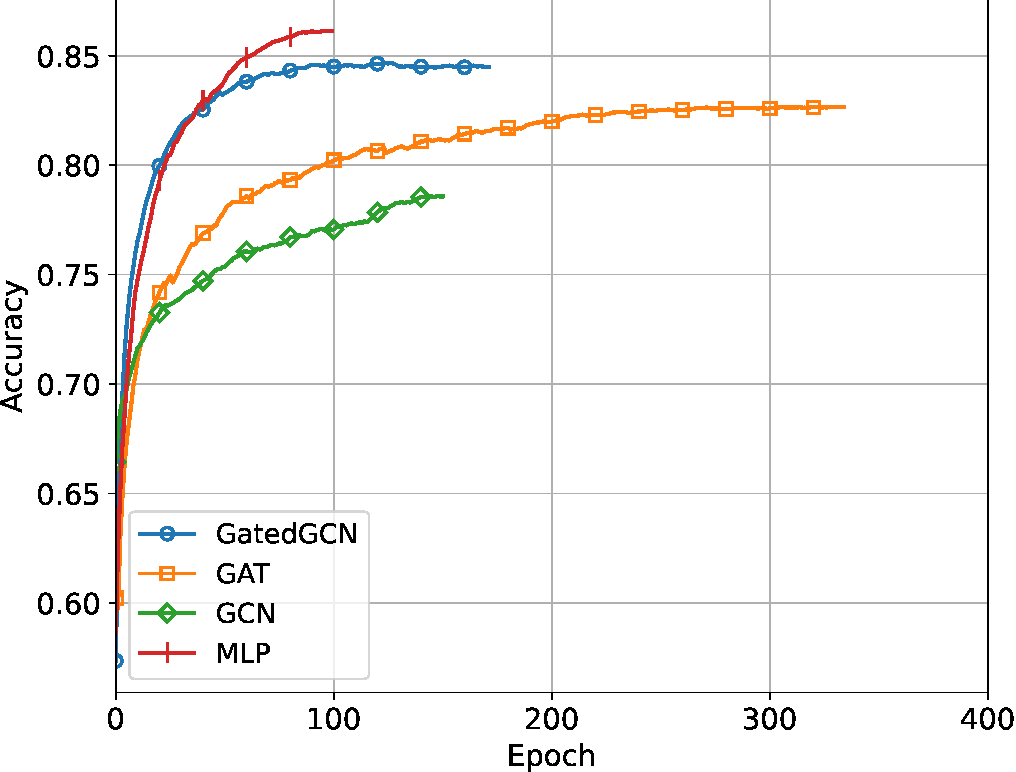}}
	\hspace{5pt}
	\subfloat[Validation accuracy in CIFAR10.]{\includegraphics[width=0.3\textwidth]{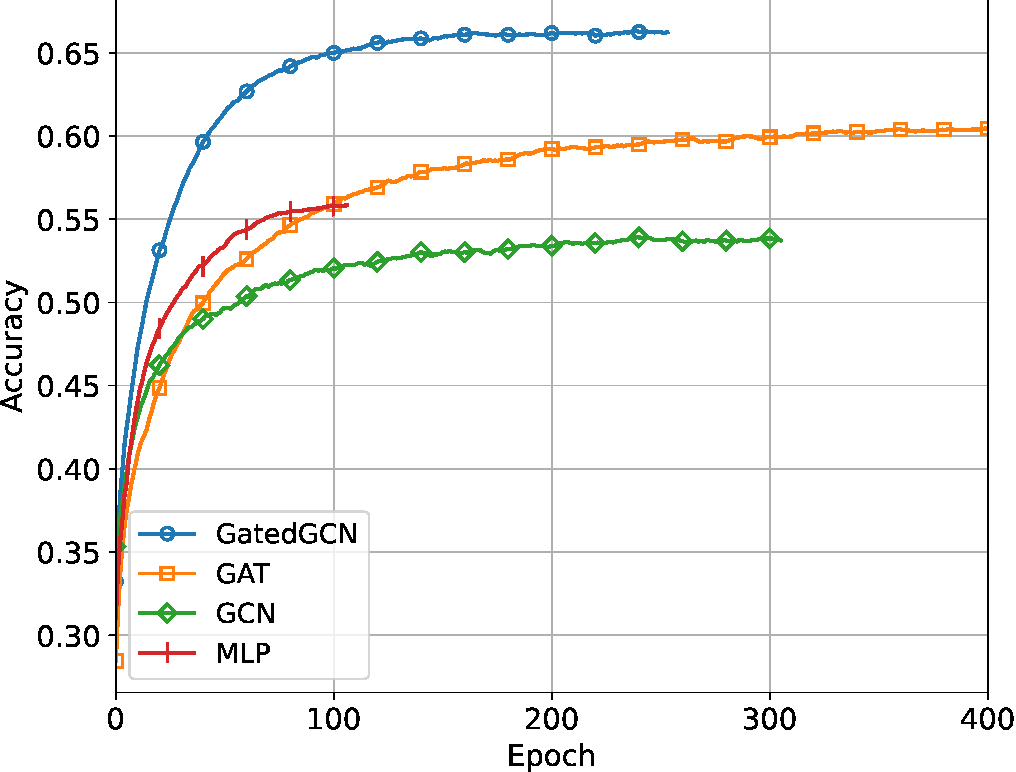}}
	\caption{Training loss and validation accuracy of GeNet with different GNN models.}
	\label{figure:train}
\end{figure*}

As for the baseline, there are only few works that are relevant to GNN-based semantic communication. We compare our proposed GeNet model with baselines as follows:
\begin{itemize}
	\item ResNet \cite{resnet}: This baseline is trained and tested on the original image dataset, serves as a benchmark representing the upper bound of GeNet's potential performance.
	\item ResNet with paint \cite{slicclassfications}: While GNN models only have access to the average colour and the centroid position, not knowing anything about the superpixel's shape. We use the paint method to simulate the color information loss of the superpixel, shown in Fig.~\ref{paint}. Specifically, we paint the original image with the average color of the superpixel. This baseline also is trained on the original image dataset but tested on the painted images.
\end{itemize}

Both ResNet20 and GNN models possess a comparable number of parameters, which is summarized in Table \ref{table:para}. In practice, the performance of ResNet20 was suboptimal on the FashionMNIST dataset. Therefore, a pre-trained ResNet18 model was utilized instead.

\begin{table}[t!]
	\renewcommand{\arraystretch}{1.3}
	\caption{Number of parameters used in different models.}
	\setlength{\tabcolsep}{4pt}
	\label{table:para}
	\begin{tabular}{cccccc}
		\toprule
		             & GatedGCN & GCN     & GAT     & MLP     & ResNet    \\ \midrule
		MNIST        & 104,217  & 101,365 & 111,008 & 105,717 & 41,610    \\
		FashionMNIST & 104,217  & 101,365 & 111,008 & 105,717 & 1,117,281 \\
		CIFAR10      & 104,357  & 101,657 & 111,312 & 106,017 & 269,722   \\ \bottomrule
	\end{tabular}
\end{table}

\subsection{Results}
\subsubsection{Train}

To assess the performance disparities of GNN models, we initially examine the GeNet model's efficacy across the MNIST, FashionMNIST, and CIFAR10 datasets with different GNN models as backbones.
Fig.~\ref{figure:train} presents the train loss and validation accuracy of the GeNet model with the configurations in Table \ref{table:config}.
% The train loss and validation accuracy of the GeNet model are shown in Fig.~\ref{figure:train}.
It's worth noting that the number of epochs trained for each model may vary slightly. This is due to the reduction of the learning rate when the validation accuracy fails to improve over several epochs, and the training process halts once the learning rate reaches a certain threshold in Section~\ref{training}.

The convergence epochs and accuracy levels demonstrate differences across the GeNet model's various backbones, owing to their distinct expressive capabilities and computational complexities. Notably, the GatedGCN model performs well on all datasets, while the MLP backbone demonstrates rapid convergence. In contrast, the GCN model exhibits subpar performance on both datasets, potentially attributed to inadequate information exchange among graph nodes.

\begin{figure*}[t!]
	\centering
	\subfloat[Test accuracy in MNIST.]{\includegraphics[width=0.3\linewidth]{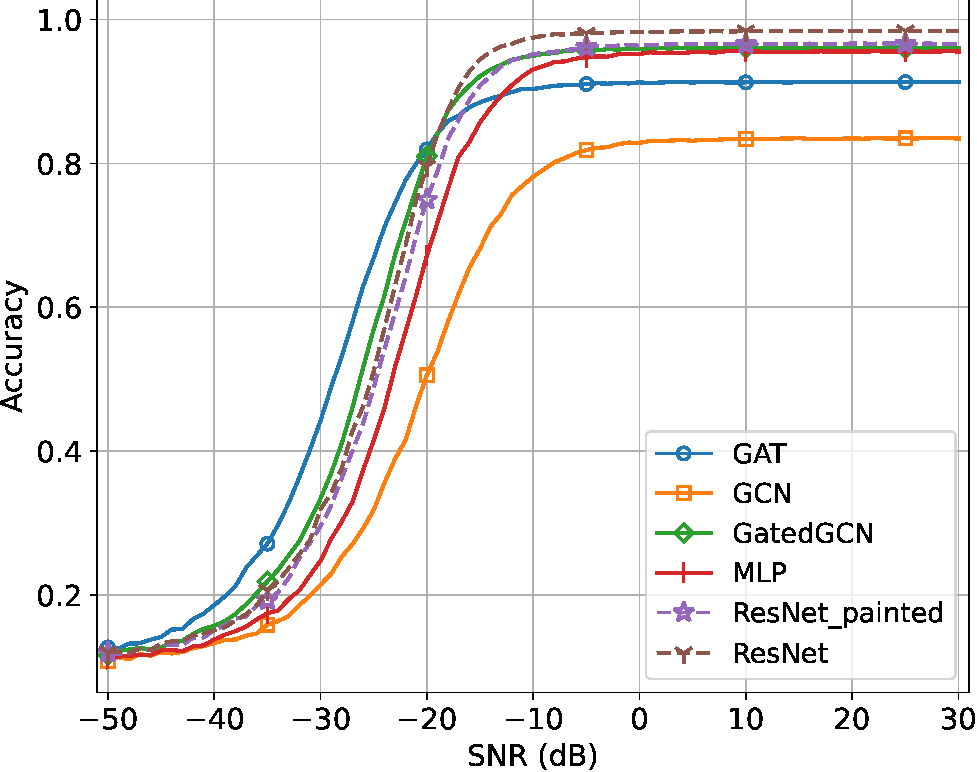}}
	% \hspace{20pt}
	\hspace{5pt}
	\subfloat[Test accuracy in FashionMNIST.]{\includegraphics[width=0.3\linewidth]{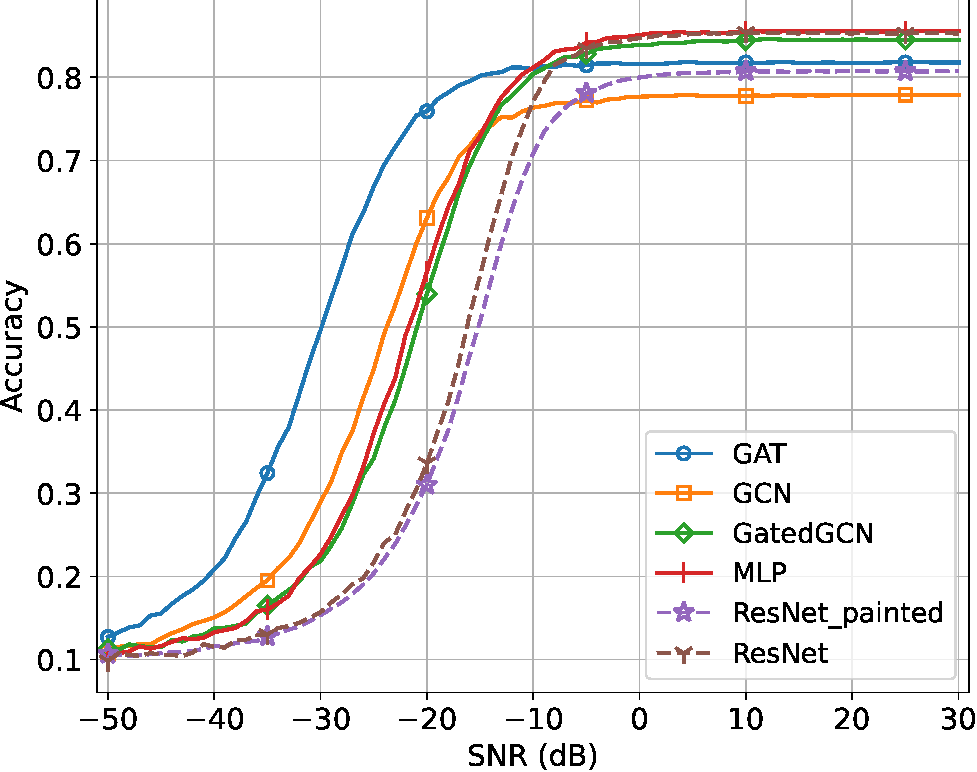}}
	\hspace{5pt}
	\subfloat[Test accuracy in CIFAR10.]{\includegraphics[width=0.3\linewidth]{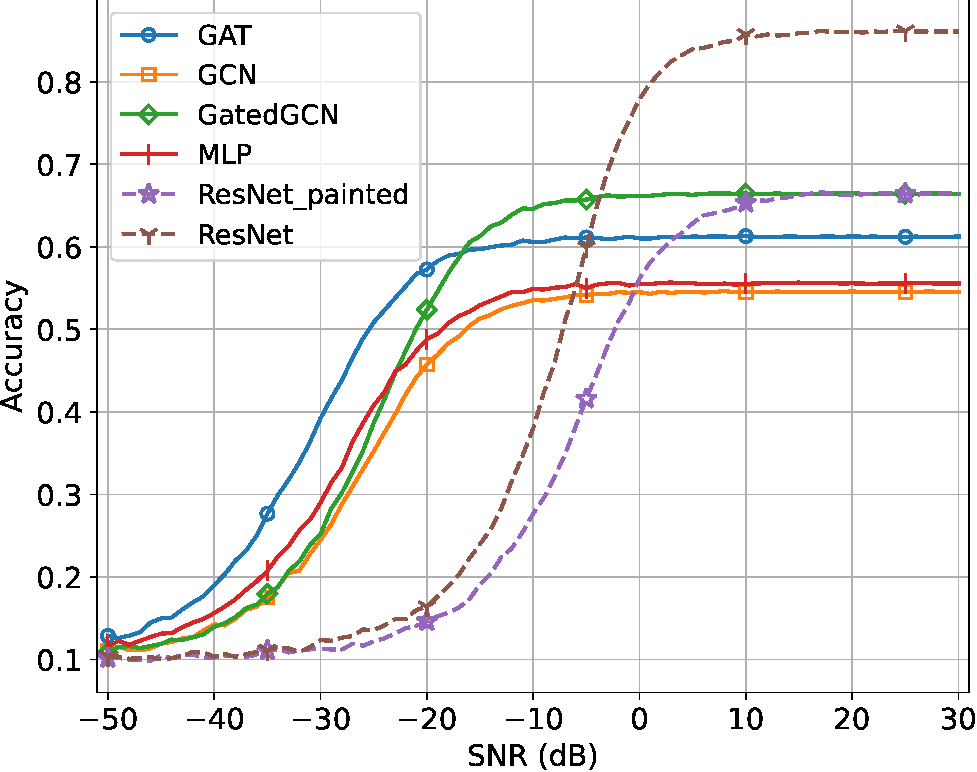}}
	\caption{Test accuracy versus SNR of different models on three test datasets.}
	\label{figure:snr}
\end{figure*}

\begin{figure*}[t!]
	\centering
	\subfloat[Test accuracy in MNIST.]{\includegraphics[width=0.3\linewidth]{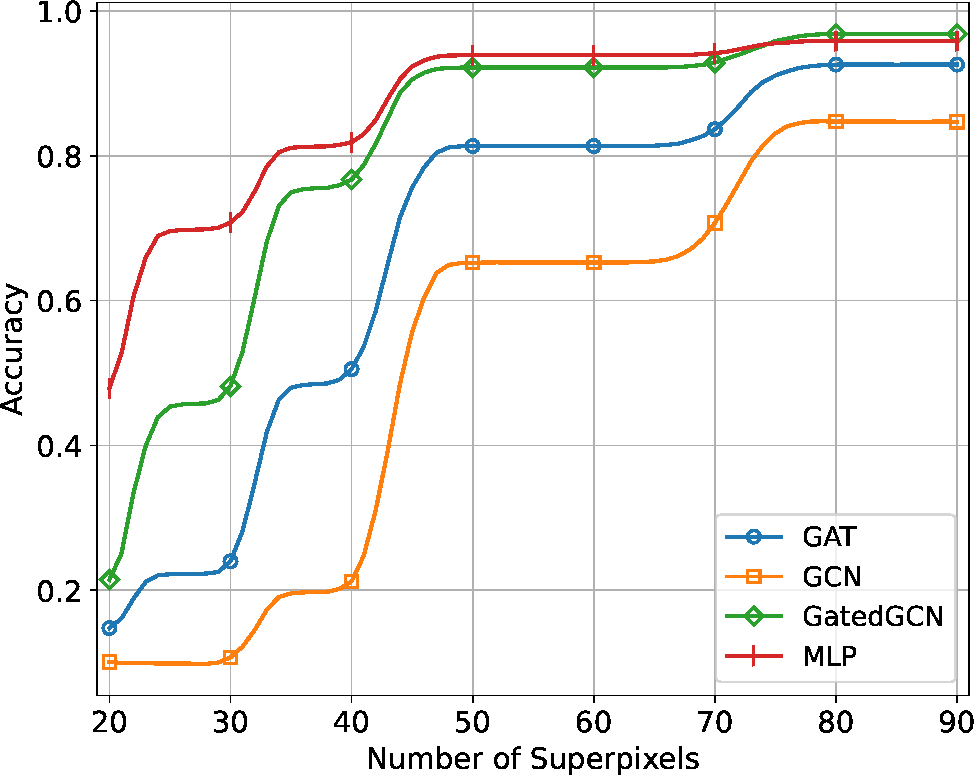}}
	\hspace{5pt}
	\subfloat[Test accuracy in FashionMNIST.]{\includegraphics[width=0.3\linewidth]{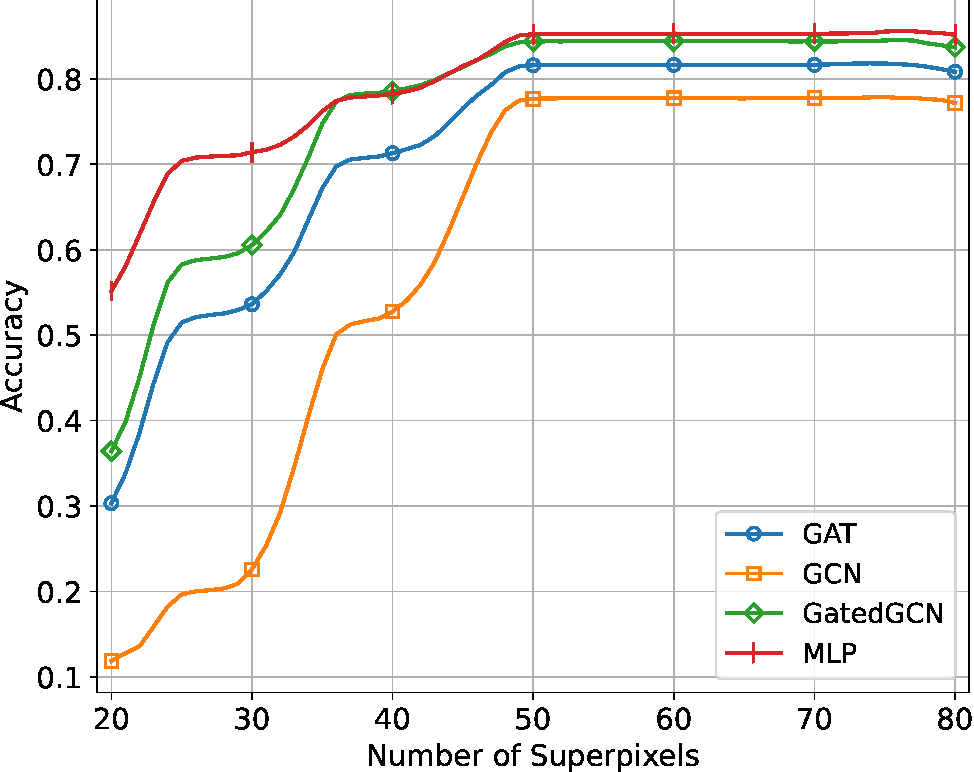}}
	\hspace{5pt}
	\subfloat[Test accuracy in CIFAR10.]{\includegraphics[width=0.3\linewidth]{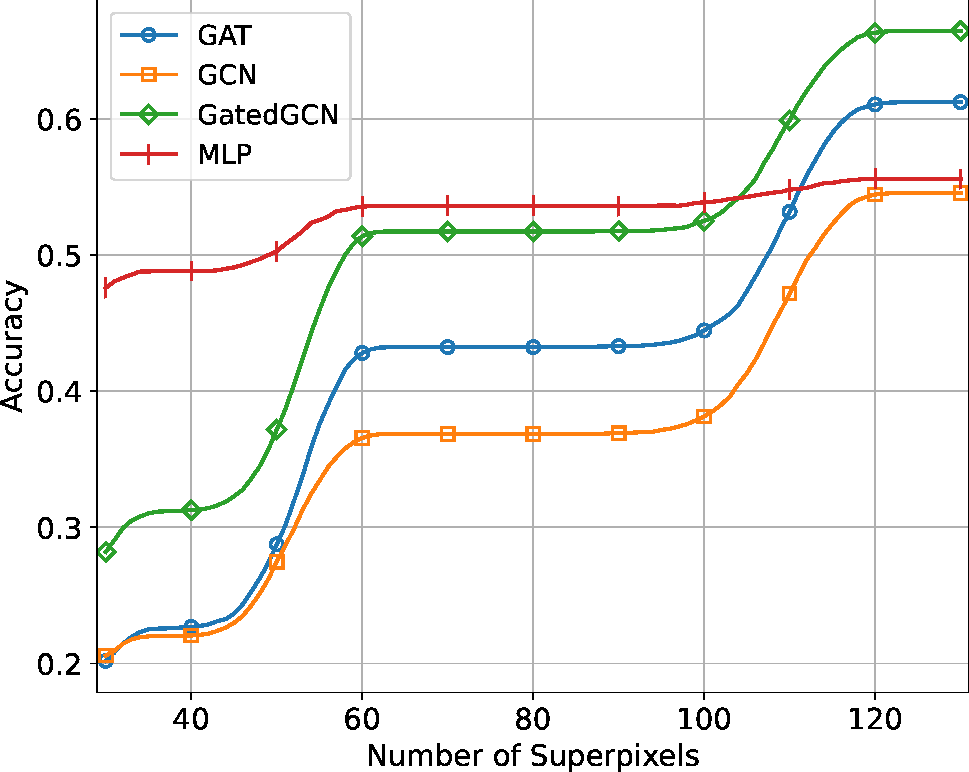}}
	\caption{Test accuracy versus number of superpixels of different models on three test datasets.}
	\label{figure:nsp}
\end{figure*}

\begin{figure*}[t!]
	\centering
	\subfloat[Test accuracy in MNIST.]{\includegraphics[width=0.3\linewidth]{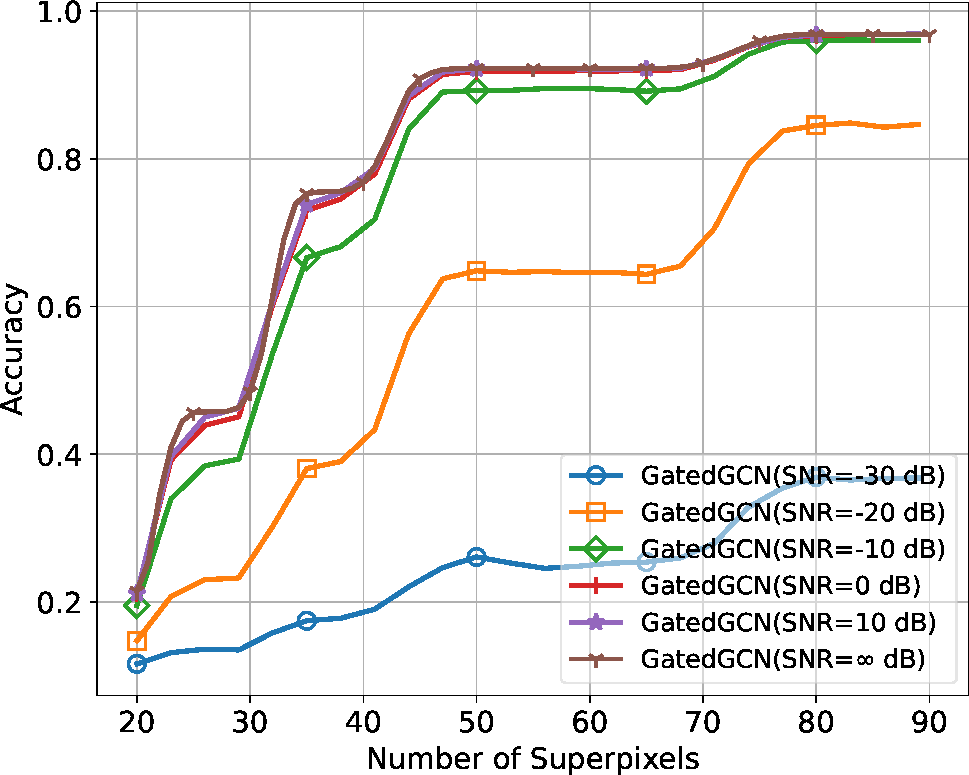}}
	\hspace{5pt}
	\subfloat[Test accuracy in FashionMNIST.]{\includegraphics[width=0.3\linewidth]{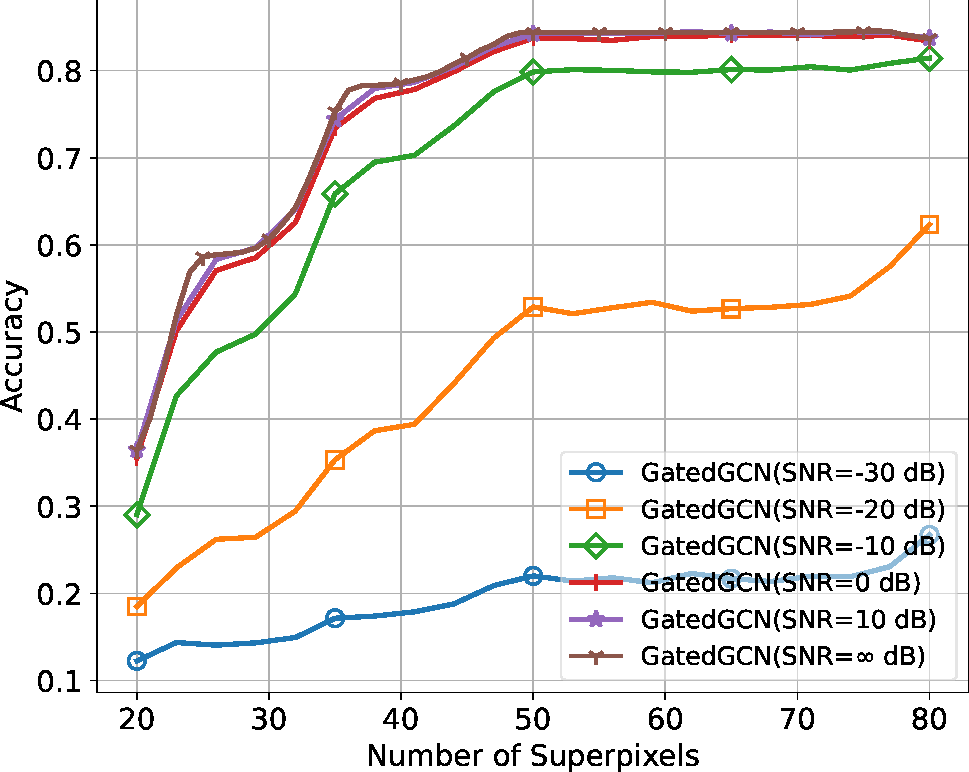}}
	\hspace{5pt}
	\subfloat[Test accuracy in CIFAR10.]{\includegraphics[width=0.3\linewidth]{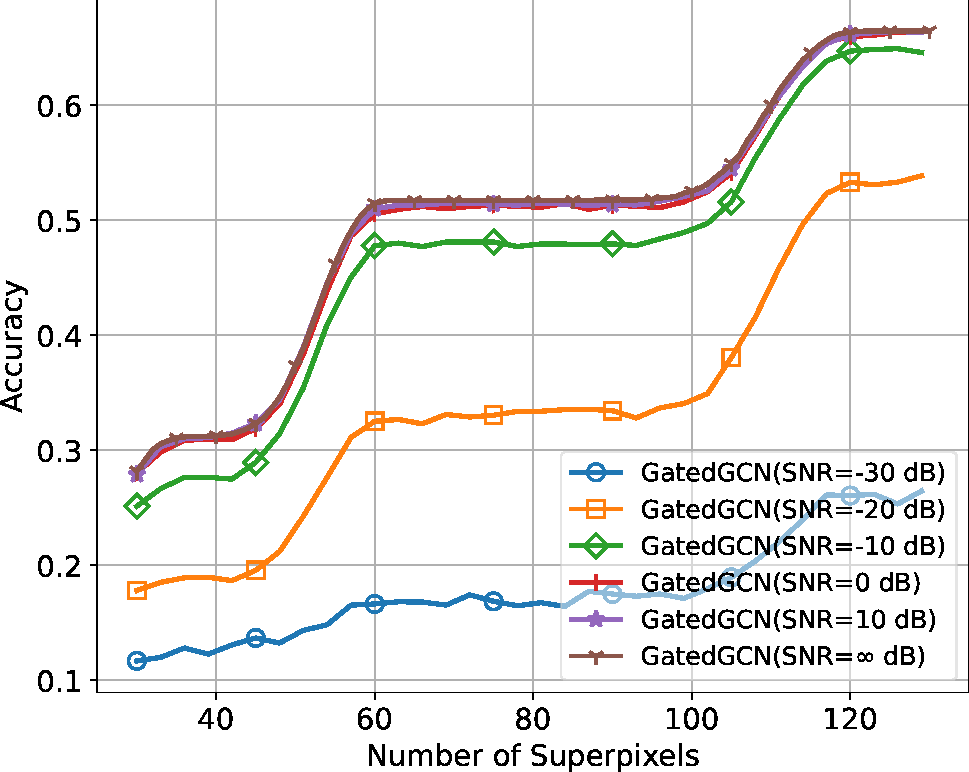}}
	\caption{Test accuracy versus number of superpixels of GatedGCN under different SNRs on three test datasets.}
	\label{figure:cross}
\end{figure*}

\subsubsection{Evaluation with SNR} \label{SNR}
We subsequently assessed the performance of the GeNet model on the test sets of MNIST, FashionMNIST and CIFAR10 datasets across varying SNR levels, aiming to ascertain its effectiveness in mitigating noise in task-oriented communication. The outcomes of this evaluation are illustrated in Fig.~\ref{figure:snr}.

Notably, our model exhibits superior performance in low SNR environments compared to baseline models.
However, as the SNR increases, compared with the ResNet model as the upper bound, the performance of the GeNet model is still not satisfactory especially in CIFAR10 dataset. This may be due to the loss of information in the superpixel graph structure. This disparity in performance could potentially be attributed to the information loss inherent in the superpixel graph structure. The GeNet shows marginal improvement in noise resistance compared to the baseline on the MNIST dataset. This outcome could be attributed to the MNIST's inherent simplicity, where the influence of noise is relatively minimal.

\subsubsection{Evaluation with number of superpixel nodes}
We further assess the performance of the GeNet model on the test sets of MNIST, FashionMNIST and CIFAR10 datasets, varying the number of superpixel nodes to explore the possibility of it serving as a semantic communication paradigm. The results are shown in Fig.~\ref{figure:nsp}. The model's performance demonstrates enhancement with an increase in the expected number of superpixel nodes.

Moreover, we also conducted the evaluation with the number of superpixel nodes for different SNR levels on the test sets. The results are shown in Fig.~\ref{figure:cross}. We notice a trend where the reduction in the number of nodes correlates with a decline in the test set accuracy. This phenomenon can be attributed to two primary factors: firstly, a lower number of nodes results in diminished extraction of semantic information; secondly, according to the equation \eqref{eq:mean_snr}, the SNR is inversely proportional to the number of nodes, and the noise level increases as the number of nodes decreases. This trend is consistent across all datasets and all GNN models. Based on this observation, we can do a trade-off in real-world communication systems by adjusting the number of superpixel nodes to balance the communication overhead and task accuracy.

Furthermore, Fig.~\ref{figure:cross} also illustrates that the performance of the model at an SNR of 0 dB is nearly identical to its performance in the absence of noise. Similarly, at an SNR of -10 dB, the model's effectiveness remains closely matched to the noise-free scenario. This indicates indirectly that the GeNet model has a certain ability to resist noise, and the performance of the model is not significantly affected by noise when the SNR is relatively low.

% \begin{figure}[t!]
%     \centering
%     \subfloat[Test accuracy (MNIST)]{\includegraphics[width=0.5\linewidth]{figure/mnist_n_sp.eps}}
%     \subfloat[Test accuracy (CIFAR10)]{\includegraphics[width=0.5\linewidth]{figure/cifar10_n_sp.eps}}
%     \caption{Evaluation with number of superpixels}
%     \label{figure:nsp}
% \end{figure}

\subsubsection{Evaluation with rotation angles}
\begin{figure}[!t]
	\centering
	\includegraphics[width=0.85\linewidth]{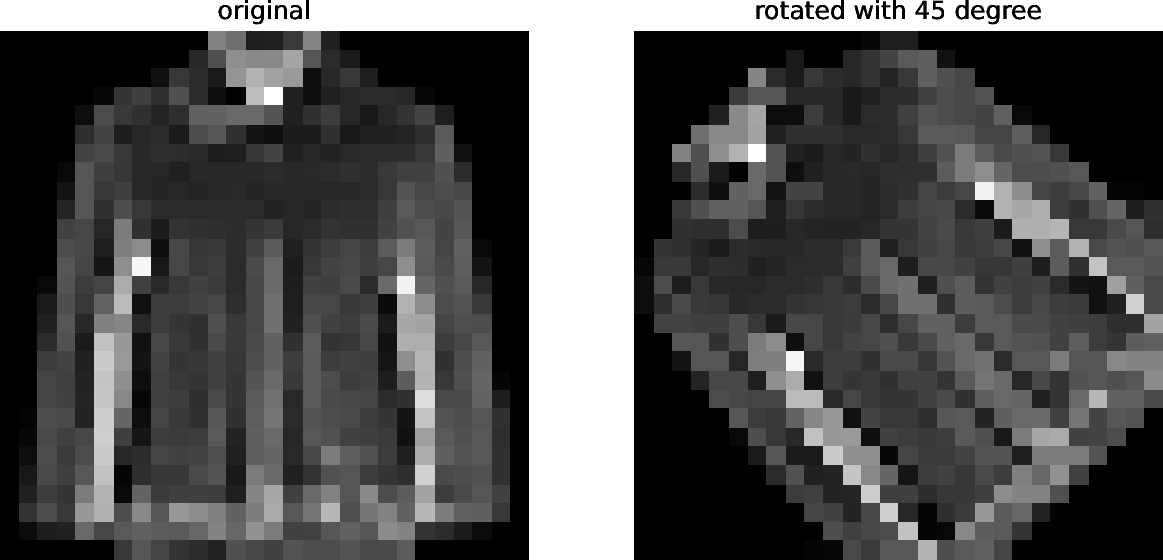}
	\caption{An example of rotated image from FashionMNIST.}
	\label{example:rotation}
\end{figure}

\begin{figure}[t!]
	\centering
	% \subfloat[Test accuracy in MNIST]{\includegraphics[width=0.45\linewidth]{figure/mnist_rotation.eps}}
	\includegraphics[width=0.8\linewidth]{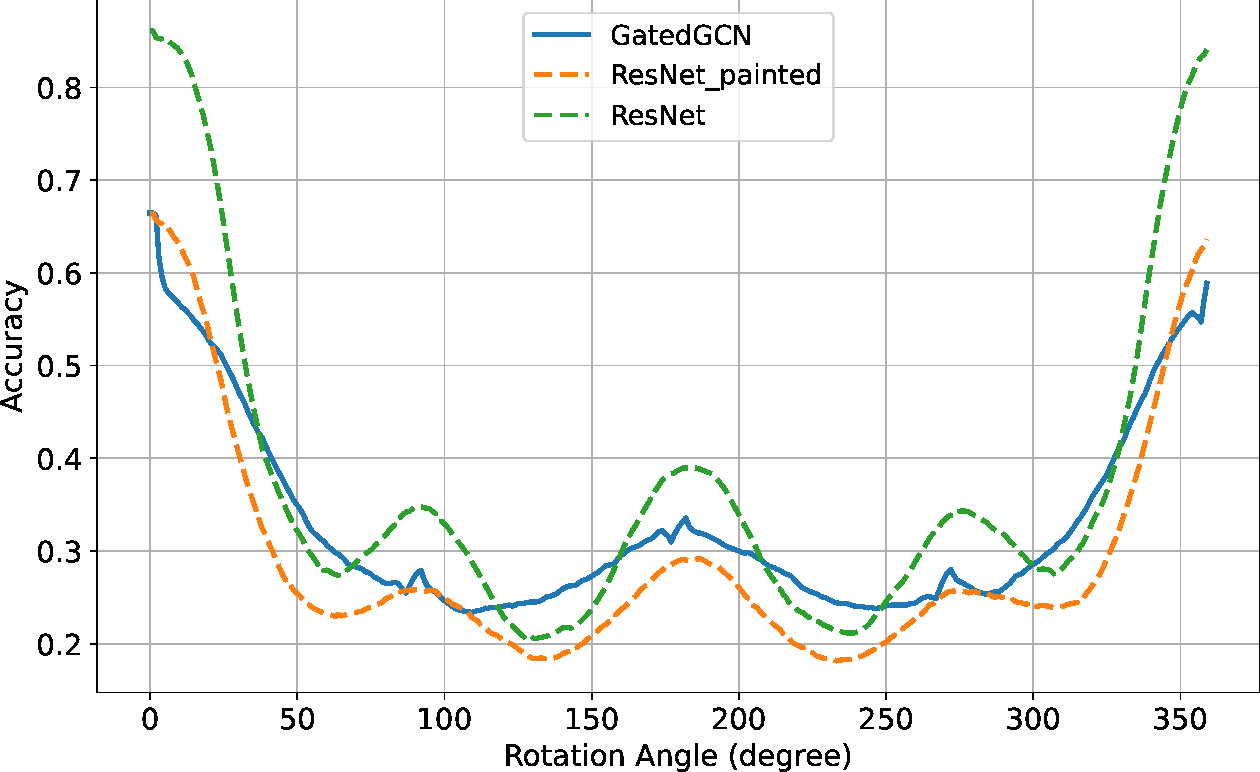}
	\caption{Test accuracy versus rotation angles of different models in CIFAR10.}
	\label{figure:rotation}
\end{figure}

Moreover, as an incidental outcome of this study, we also explored task-oriented communication systems for images with geometric transformations. This facet allows the communication system to accurately infer tasks even when confronted with image data subjected to geometric alterations like rotation or translation which is shown in Fig.~\ref{example:rotation}.
To substantiate this capability, we conducted evaluations of the GeNet model on test sets of MNIST, FashionMNIST and CIFAR10 datasets, varying the rotation angles. The results are similar for all datasets, only the CIFAR10 dataset is shown in Fig.~\ref{figure:rotation}.

When the rotation angle aligns precisely with an integer multiple of 90 degrees, a noticeable improvement in accuracy on the test set is observed. This enhancement can be linked to the fact that, under these particular conditions, the images post-rotation maintain a greater level of regularity and endure the least amount of information loss compared to rotations at other angles.

Like the evalution with SNR in Section \ref{SNR}, our model exhibits superior performance compared to painted baseline models, but inferior to the ResNet model as the upper bound for some rotation angles due to the loss of information.
The findings conclusively illustrate the GeNet model's proficiency in handling such transformations, showcasing its robustness and adaptability across diverse scenarios without data augmentation.

\section{Conclusion}\label{Conclusion}

In this paper, we introduce GeNet, a novel paradigm leveraging Graph Neural
Networks (GNNs) for anti-noise, task-oriented semantic communications.
Different from traditional methods, GeNet transforms input data images into
graph structures and utilizes a GNN-based encoder-decoder model to extract and
reconstruct semantic information, respectively.
Through extensive experiments on MNIST, FashionMNIST, and CIFAR10 datasets, we
have demonstrated that our model is superior performance in anti-noise TOC
without prior knowledge of channel conditions, thus saving time and
computational resources required for traditional methods trained at different
SNRs.
Moreover, GeNet can process images of varying resolutions without resizing,
ensuring preservation of information integrity, and exhibits robustness to
geometric transformations such as rotations without data augmentation.
Moving forward, exploring GeNet's potential in transmitting data originally
structured as graphs, handling images of different sizes, addressing geometric
transformations, and refining graph transformation and noise handling schemes
are promising avenues for further research in semantic communication.

\section*{Acknowledgements}

This work was supported in part by National Key R\&D
Program of China under Grant 2022YFB2902700, NSF China (Grant No. 
62202508, 62071501), and Shenzhen Science and Technology Program (Grant
20220817094427001, JCYJ20220818102011023, ZDSYS20210623091807023).

\bibliographystyle{IEEEtran}

%%%%% CLEAR DOUBLE PAGE!
\newpage{\pagestyle{empty}\cleardoublepage}

\end{document}